%% file: main.tex
\documentclass{article}
\usepackage{spconf,amsmath,graphicx}
\usepackage{hyperref}
\usepackage{url}
\usepackage{graphicx}
\usepackage{subcaption}
\graphicspath{{figures/}}
\input{macrosjvg}

\title{Exploiting Learned Symmetries in Group Equivariant Convolutions}

\name{Attila Lengyel \qquad Jan van Gemert\thanks{This project is supported in part by NWO (project VI.Vidi.192.100).}}
\address{Computer Vision Lab, Delft University of Technology, The Netherlands}

\begin{document}

%\ninept
\maketitle

\begin{abstract}
Group Equivariant Convolutions (GConvs) enable convolutional neural networks to be equivariant to various transformation groups, but at an additional parameter and compute cost. We investigate the filter parameters learned by GConvs and find certain conditions under which they become highly redundant. We show that GConvs can be efficiently decomposed into depthwise separable convolutions while preserving equivariance properties and demonstrate improved performance and data efficiency on two datasets. All code is publicly available at \textcolor{blue}{github.com/Attila94/SepGrouPy}.
\end{abstract}

\begin{keywords}
group equivariant convolutions, depthwise separable convolutions, efficient deep learning
\end{keywords}

\input{sections/01_introduction}
\input{sections/02_related_work}
\input{sections/03_method}
\input{sections/04_experiments}
\input{sections/05_discussion}

%\vfill
%\pagebreak

\bibliographystyle{IEEEbib}
\bibliography{refs}

\end{document}

%% file: macrosjvg.tex
\usepackage{xcolor}

\newcommand{\Eqs}[2]{Eqs.~\ref{eq:#1}-\ref{eq:#2}}
\newcommand{\Eq}[1]{Eq.~(\ref{eq:#1})}
\newcommand{\eq}[1]{\Eq{#1}}
\newcommand{\fig}[1]{Fig.~\ref{fig:#1}}
\newcommand{\tab}[1]{Table~\ref{tab:#1}}

\usepackage{pifont}
\usepackage{tabularx} % extra features for tabular environment
\newcolumntype{Y}{>{\centering\arraybackslash}X}

\usepackage{booktabs}
\usepackage{multirow}
\usepackage{tablefootnote}

%% file: sections/01_introduction.tex
\section{Introduction}
\label{sec:intro}

Adding convolution to neural networks (CNNs) yields translation equivariance~\cite{kayhan2020translation}: first translating an image $x$ and then convolving is the same as first convolving $x$ and then translating. %Translation equivariance allows parameter sharing over the spatial input dimensions, thereby greatly reducing the amount of required training data since the model does not need to learn the object in all different locations. 
Group Equivariant Convolutions~\cite{cohen2016group} (GConvs) enable equivariance to a larger group of transformations $G$, including translations, rotations of multiples of 90 degrees ($p4$ group), and horizontal and vertical flips ($p4m$ group). Equivariance to a group of transformations $G$ is guaranteed by sharing parameters between  filter copies for each transformation in the group $G$. Adding such geometric symmetries as prior knowledge offers a hard generalization guarantee to all transformations in the group, reducing the need for large annotated datasets and extensive data augmentation.

\begin{figure}[t]
    \centering
    \includegraphics[width=\linewidth]{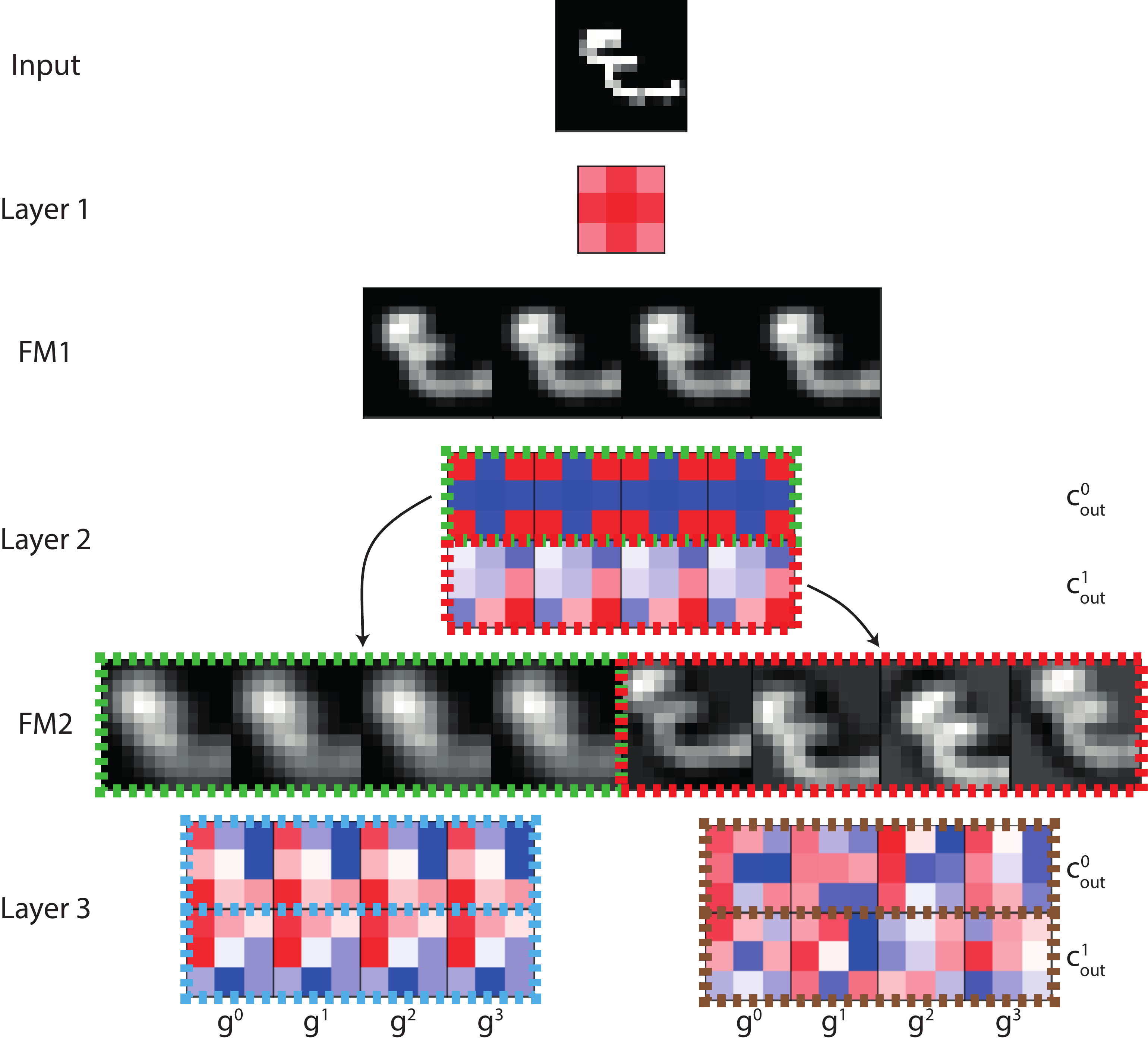}
    \caption{Filters and feature maps of a GConv architecture trained on Rotated MNIST. Rotation invariant filters in Layer 2 result in identical feature maps FM2 (green) and cause Layer 3 to learn identical weights along the group dimension $g$ (blue). In contrast, non-symmetric filters in Layer 2 (red) result in non-identical filters in Layer 3 (brown).}
    \label{fig:toyexp}
\end{figure}

In practice, however, GConvs occasionally learn filters that are near-invariant to transformations in $G$. An invariant filter is independent of the transformation and will for GConvs yield identical copies of the transformed filters in the consecutive layer, as shown in~\fig{toyexp}. This implies parameter redundancy, as these filters could be represented by a single spatial kernel. We propose an equivariant pointwise and a depthwise decomposition of GConvs with increased parameter sharing and thus improved data efficiency. Motivated by the observed inter-channel correlations in learned filters in~\cite{Haase_2020_CVPR} we explore additionally sharing the same spatial kernel over all input channels of a GConv filter bank. Our contributions are: (i) we show that near-invariant filters in GConvs yield  highly correlated spatial filters; (ii) we derive two decomposed GConv variants; and (iii) improve accuracy compared to GConvs on RotMNIST and CIFAR10.

%% file: sections/02_related_work.tex
\section{Related Work}
\textbf{Equivariance in deep learning.}
Equivariance is a promising research direction for improving data efficiency~\cite{Rath2020BoostingDN}. A variety of methods have extended the Group Equivariant Convolution for the $p4$ and $p4m$ groups introduced in~\cite{cohen2016group} to larger symmetry groups including translations and discrete 2D rotations~\cite{bekkers2018,Weiler_2018_CVPR}, 3D~rotations~\cite{winkels2019,worrall2018,weiler2018a}, and scale~\cite{worrall2019,Sosnovik2020Scale-Equivariant}. Here, we investigate learned invariances in the initial GConv framework~\cite{cohen2016group} for the $p4$ and $p4m$ groups, yet our analysis extends to other groups where invariant filters exist.
% including the $\mathbf{SE}(2,N)$ group of translations and $N$ 2D rotations~\cite{}, the $\mathbf{SE}(2)$ and $\mathbf{SE}(3)$ groups of translations and arbitrary 2D~\cite{} and 3D rotations~\cite{}, respectively, and the $S$ group of scale~\cite{}. In this work we investigate the learned invariances in the initial GConv framework~\cite{cohen2016group} for the $p4$ and $p4m$ groups, yet our analysis extends readily to other groups.

\noindent \textbf{Depthwise separable decomposition~\cite{Sifre14}.} These decompose a multi-channel convolution into spatial convolutions applied on each individual input channel separately, followed by a pointwise (1x1) convolution. Depthwise separable convolutions significantly reduce parameter count and computation cost at the expense of a slight loss in representation power and therefore generally form the basis of network architectures optimized for efficiency~\cite{Howard2019SearchingFM, tan2019, mohamed2020data}. The  effectiveness of depthwise separable convolutions is motivated~\cite{Haase_2020_CVPR} by the observed inter-channel correlations occurring in the learned filter banks of a CNN, which is quantified using a PCA decomposition. We do a similar analysis to motivate and derive our separable implementation of GConvs. 

% Deze zin zou ik pas in de camera-ready opnemen: (paper wel citeren)
%While such implementation is briefly mentioned in \cite{mohamed2020data}, it lacks any explanation or analysis provided in our work.

%% file: sections/03_method.tex
\section{Method}
\subsection{Group Equivariant Convolutions}
Equivariance to a group of transformations $G$ is defined as
\begin{align}
    \Phi(T_gx) = T'_g\Phi(x), \quad \forall g \in G,
\end{align}
where $\Phi$ denotes a network layer and $T_g$ and $T'_g$ a transformation $g$  on the input and feature map, respectively. Note that in the case of translation equivariance $T$ and $T'$ are the same, but in general not need to be. To simplify the explanation, we first focus on the group $p4$ of translations and 90-degree rotations, but extend to larger groups later. 

Let us denote a regular convolution as
\begin{align}
\label{eq:conv}
    X_{n,:,:}^{l+1} = \sum_c^{C^{l}} F^l_{n,c,:,:} * X_{c,:,:}^l,
\end{align}
with $X$ the input and output tensors of size $[C^l,H,W]$, where $C^l$ is the number of channels in layer $l$, H is height and W is width, and $F$ the filter bank of size $[C^{l+1},C^{l},k,k]$, with $k$ the spatial extent of the filter.

% GConvs differ by adding a filter dimension for the added transformation group, which for added rotations yields a bank $F^l$ of size $[C^{l+1},C^{l},G^{l},k,k]$, where $G^{l}$ denotes the size of the rotation group $G$ at layer $l$. As such, filter banks in GConvs contain $G^l$ times more trainable parameters compared to regular convolutions. A full GConv \jvg{May be unclear; what is 'full'? It seems to say that you have 2 versions of a GConv filter bank (a 'normal' one, and a 'full' one?). I understand you first want to talk about the added group size, but perhaps this can be done by explaining how $X$ changes for GConvs vs normal convs? (as the change in $X$ in \eq{gconv} vs \eq{conv} is never mentioned.} filter $\Tilde{F}^l$ of size $[C^{l+1},G^{l+1},C^{l},G^{l},k,k]$ is constructed during each forward pass, where the additional dimension $G^{l+1}$ contains rotated and cyclically permuted versions of $F^l$~\cite{cohen2016group}. A GConv is then performed as

In addition to spatial location, GConvs  encode the added transformation group $G$ in an extra tensor dimension such that $X$ becomes of size $[C^l,G^l,H,W]$, where $G^{l}$ denotes the size of the transformation group $G$ at layer $l$, i.e. 4 for the $p4$ group. Likewise, GConv filters acting on these feature maps contain an additional group dimension, yielding a filter bank $F^l$ of size $[C^{l+1},C^{l},G^{l},k,k]$. As such, filter banks in GConvs contain $G^l$ times more trainable parameters compared to regular convolutions. A GConv is then performed by convolving over both the input channel and input group dimensions $C^l$ and $G^l$ and summing up the outputs:
\begin{align}
\label{eq:gconv}
    X_{n,h,:,:}^{l+1} &= \sum_c^{C^{l}} \sum_g^{G^l} \Tilde{F}^l_{n,h,c,g,:,:} * X_{c,g,:,:}^l.
\end{align}
Here $\Tilde{F}^l$ denotes the full GConv filter of size $[C^{l+1}, G^{l+1},\allowbreak C^{l},G^{l},k,k]$ containing an additional dimension for the output group $G^{l+1}$. $\Tilde{F}^l$ is constructed from $F^l$ during each forward pass, where $G^{l+1}$ contains rotated and cyclically permuted versions of $F^l$ (see ~\cite{cohen2016group} for details). Note that input images do not have a group dimension, so the input layer has $G^l{=}1$ and $X^1_{c,g,:,:}$ reduces to $X^1_{c,:,:}$, whereas for all following layers $G^l{=}4$ for the $p4$ group (and $G^l{=}8$ for $p4m$).

% \begin{figure}
%     \centering
%      \begin{subfigure}[b]{0.4\textwidth}
%          \centering
%          \includegraphics[width=0.8\textwidth]{example-image-a}
%          \caption{GConvs}
%          \label{fig:gconvs}
%      \end{subfigure}
%      \begin{subfigure}[b]{0.4\textwidth}
%          \centering
%          \includegraphics[width=0.8\textwidth]{example-image-b}
%          \caption{Separable GConvs}
%          \label{fig:separable_gconvs}
%      \end{subfigure}
%      \caption{Overview of the method.}
% \end{figure}

\subsection{Filter redundancies in GConvs}
% Rotational symmetric filters are invariant to the relative orientation between the filter and its input. Thus, a rotational symmetric filter in a $p4$ equivariant GConv layer would produce identical feature maps along the group dimension. As a result, the filters acting on these feature maps receive identical gradients and, given same initialization, learn identical filters.
A rotational symmetric filter is invariant to the relative orientation between the filter and its input. Thus, if the filter kernels in the group dimension of a $p4$ GConv filter bank $F^l$ are rotational symmetric and identical, the resulting feature maps will also be identical along the group dimension due to the rotation and cyclic permutation performed in constructing the full filter bank $\Tilde{F}^l$. As a result, the filters in the subsequent layer acting on these feature maps receive identical gradients and, given same initialization, learn identical filters. This is illustrated in \fig{toyexp} where a $p4$ equivariant CNN is trained on Rotated MNIST. The first layer contains a single fixed rotation invariant filter. All layers have equal initialization along the group dimension and linear activation functions. The filters in layer 2 converge to be identical along the group dimension. Furthermore, the filter kernels in the second layer belonging to the first output channel (green) are also rotational symmetric, resulting in identical feature maps in FM2 (green) and consequently the filters learned in the first input channel of layer 3 (blue) become highly similar. This is in contrast to the non-symmetric filters in layer 2 (red), resulting in non-identical filters in layer 3 (brown).
% , whereas the filters acting on the edge detector feature maps are inversely correlated (i.e. $g_0 \approx -g_2$ and $g_1 \approx -g_3$)

Even non-rotational symmetric filters can induce filter correlations in the subsequent layer. For instance, an edge detector will result in inverse feature maps along the group dimension, i.e. $g^0{\approx}-g^2$ and $g^1{\approx}-g^3$ and the filters acting on these feature maps will receive inverse gradients and consequently converge to be inversely correlated. Inversely correlated filters can be decomposed into the same spatial kernel multiplied by a positive and negative scalar.

\begin{figure*}
    \centering
    \includegraphics[width=\textwidth]{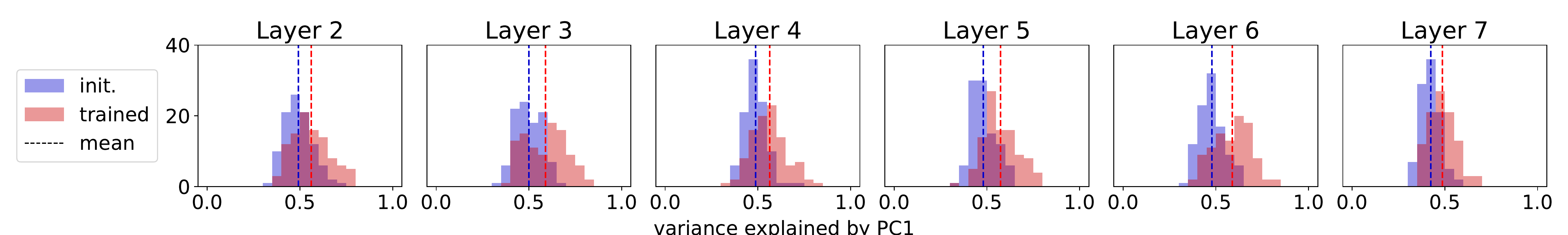}
    \caption{Ratio of variance explained by the first principal component when decomposing a filter kernel along the group dimension, before (blue) and after (red) training on Rotated MNIST. Redundancy in filter parameters increases as the network converges.}
    \label{fig:hist}
\end{figure*}

% \jvg{We quantify how often GConv filters correlate along the group dimension.}
Upon visual inspection of the learned filter parameters of a regular $p4$ equivariant CNN we observe that, even without any fixed symmetries or initialization and with ReLU activation functions, the filter kernels tend to be correlated along the group axis. To quantify this correlation we perform a PCA decomposition similar as in \cite{Haase_2020_CVPR}. We reshape the filter bank $F$ to size $[C^{l+1}\times C^{l},G^{l},k^2]$ and perform PCA on each set of filters $F_{n,:,:}$ for all $n \in [1,C^{l+1}\times C^{l}]$, where for each $n$ we have $G^l$ features with $k^2$ samples. This results in $G^l$ principal components of size $k^2$, with PC1 being the filter kernel explaining the most variance within the decomposed set.
% This is illustrated in \fig{p4_filterbank}, where filterbank $F_{n,:,:,:,:}$ for output channel $n$ of a $p4$ equivariant CNN is shown, with the input channels on the x-axis and the input group dimension on the y-axis (rows 2-5). The filters in several columns are clearly (inversely) correlated. The PC1 filter kernel resulting from decomposing each column is shown in the top row. The score denotes the fraction of variance explained by PC1. 
We perform this decomposition for all layers in a $p4$ equivariant network. \fig{hist} shows the ratio of the variance explained by PC1 for each layer (after the input layer), before and after training. In many cases a substantial part of the variance is explained by a single component, demonstrating a significant redundancy in filter parameters.
% On average 56\% of the variance is explained by a single component, demonstrating a significant redundancy in filter parameters.
% \begin{figure}
% \centering
% \includegraphics[width=0.8\linewidth]{figures/p4_filterbank.pdf}
% \caption{Filterbank $F_{n,:,:,:,:}$ for output channel $n$ of a $p4$ equivariant CNN. The filters are often correlated in the group dimension (columns). Top row shows the first principal component resulting from a PCA decomposition of the four filters below and the fraction of variance explained by PC1.}
% \label{fig:p4_filterbank}
% \end{figure}

\subsection{Separable Group Equivariant Convolutions}

To exploit the correlations in GConvs we decompose the filter bank $F^l$ into a 2D kernel $K$ that is shared along the group dimension, and a pointwise component $w$ which encodes the inter-group correlations:
\begin{align}
    F^l_{n,c,g,:,:} &= K^l_{n,c,:,:} \cdot w^l_{n,c,g}.
\end{align}
The full GConv filter bank is then constructed as
\begin{align}
\label{eq:filter_decomp}
    \Tilde{F}^l_{n,h,c,g,:,:} &= T_h(K^l_{n,c,:,:}) \cdot \Tilde{w}^l_{n,h,c,g},
\end{align}
where $T_h$ denotes the 2D transformation corresponding to output group channel $h$ and $\Tilde{w}^l$ contains copies of $w^l$ that are cyclically permuted along the input group dimension. A naive implementation would be to precompute $\Tilde{F}$ and perform a regular GConv as in \eq{gconv}. Alternatively, for better computational efficiency we can substitute the filter decomposition in \eq{filter_decomp} into the GConv in \eq{gconv} and rearrange as follows:
\begin{align}
    X_{n,h,:,:}^{l+1} &= \sum_c^{C^{l}} \sum_g^{G^l} X^l_{c,g,:,:} * \left( T_h(K^l_{n,c,:,:}) \cdot \Tilde{w}^l_{n,h,c,g} \right) \\
    &= \sum_c^{C^{l}} \sum_g^{G^l} \left(X^l_{c,g,:,:} \cdot \Tilde{w}^l_{n,h,c,g} \right) *  T_h(K^l_{n,c,:,:})  \\
    \label{eq:gconv_g_sep}
    &= \sum_c^{C^{l}} \Tilde{X}^l_{n,h,c,:,:} *  T_h(K^l_{n,c,:,:})
\end{align}
with
\begin{align}
\label{eq:pointwise_g}
    \Tilde{X}^l_{n,h,c,:,:} = \sum_g^{G^l} \left(X_{c,g,:,:}^l \cdot \Tilde{w}^l_{n,h,c,g} \right).
\end{align}
Expanding the dimensions of $\Tilde{w}^l$ to $[C^{l+1},G^{l+1},C^{l},G^{l},1,1]$ we can implement \eq{pointwise_g} as a grouped $1\times1$ convolution with $C^l$ groups, followed by a grouped spatial convolution with $C^{l+1}\times G^{l+1}$ groups, as given in \eq{gconv_g_sep}. We refer to this separable GConv variants as $g$-GConv, denoting the summation variable in \eq{pointwise_g}.

Alternatively, we share the spatial kernel $K$ along both the group and input channel dimension by decomposing $F^l$ as:
\begin{align}
    F^l_{n,c,g,:,:} &= K^l_{n,:,:} \cdot w^l_{n,c,g},\\
    \Tilde{F}^l_{n,h,c,g,:,:} &= T_h(K^l_{n,:,:}) \cdot \Tilde{w}^l_{n,h,c,g}.
\end{align}
Substituting $\Tilde{F}^l$ in \eq{gconv} and rearranging yields
\begin{align}
    X_{n,h,:,:}^{l+1} &= \sum_c^{C^{l}} \sum_g^{G^l} X_{c,g,:,:}^l * \left( T_h(K^l_{n,:,:}) \cdot \Tilde{w}^l_{n,h,c,g} \right) \\
    &= \sum_c^{C^{l}} \sum_g^{G^l} \left(X_{c,g,:,:}^l \cdot \Tilde{w}^l_{n,h,c,g} \right) *  T_h(K_{n,:,:})  \\
    \label{eq:gconv_gi_sep}
    &= \Tilde{X}^l_{n,h,:,:} *  T_h(K^l_{n,:,:})
\end{align}
with
\begin{align}
\label{eq:pointwise_gi}
    \Tilde{X}^l_{n,h,:,:} = \sum_c^{C^{l}}\sum_g^{G^l} \left(X_{c,g,:,:}^l \cdot \Tilde{w}^l_{n,h,c,g} \right).
\end{align}
This way the GConv essentially reduces to an inverse depthwise separable convolution with \eq{pointwise_gi} being the pointwise and \eq{gconv_gi_sep} being the depthwise component. This variant is named $gc$-GConv after the summation variables in \eq{pointwise_gi}.

While the $g$ and $gc$ decompositions may impose too stringent restrictions on the hypothesis space of the model, the improved parameter efficiency, as detailed in section \ref{sec:compute}, allows us to increase the network width given the same parameter budget resulting in better overall performance.
% We refer to these two separable GConv variants as $g$- and $gc$-GConv, respectively, denoting the summation variables in \eq{pointwise_g} and \eq{pointwise_gi}. While such decompositions may impose too stringent restrictions on the hypothesis space of the model, the improved parameter efficiency, as detailed in section \ref{sec:compute}, allows us to increase the network width given the same parameter budget resulting in better overall performance.

\subsection{Computation efficiency} 
\label{sec:compute} 
The decomposition of GConvs allows for a theoretically more efficient implementation, both in terms of the number of stored parameters and multiply-accumulate operations (MACs). As opposed to the $[C^l\times G^l \times k^2 \times C^{l+1}]$ parameters in a GConv filter bank, $g$- and $gc$-GConvs require only $[C^l \times C^{l+1} \times (G^l + k^2)]$ and $[C^{l+1} \times (C^l \times G^l + k^2)]$, respectively. Similarly, a regular GConv layer performs $[C^l \times G^l \times k^2 \times W \times H \times C^{l+1} \times G^{l+1}]$ MACs, whereas $g$- and $gc$-GConvs do only $[C^l \times C^{l+1} \times G^{l+1} \times W \times H \times (G^l + k^2)]$ and $[C^{l+1} \times G^{l+1} \times W \times H \times (C^l \times G^l + k^2)]$, assuming 'same' padding. This translates to a reduction by a factor of $\frac{1}{k^2}+\frac{1}{G^l}$ and $\frac{1}{k^2}+\frac{1}{C^l \times G^l}$, both in terms of parameters and MAC operations. The decrease in MACs comes at the cost of a larger GPU memory footprint due to the need of storing intermediate feature maps, as is generally the case for separable convolutions. Separable GConvs are therefore especially suitable for applications where the available processing power is the bottleneck as opposed to memory.

%% file: sections/04_experiments.tex
\section{Experiments}
\subsection{Rotated MNIST}
We construct a $g$-separable (\Eqs{gconv_g_sep}{pointwise_g}) and $gc$-separable (\Eqs{gconv_gi_sep}{pointwise_gi}) version of the P4CNN architecture~\cite{cohen2016group} and evaluate on Rotated MNIST~\cite{larochelle2007}. Rotated MNIST has 10 classes  of randomly rotated handwritten digits with 12k train and 60k test samples. We set the width $w$ of the $g$-P4CNN and $gc$-P4CNN networks such that the number of parameters are as close as possible to our Z2CNN and P4CNN baselines of 10 and 20 channels, respectively. We follow the training procedure of~\cite{cohen2016group} and successfully reproduced the results.

\begin{figure}
    \centering
    \begin{subfigure}[b]{0.50\linewidth}
        \centering
        \includegraphics[width=\textwidth]{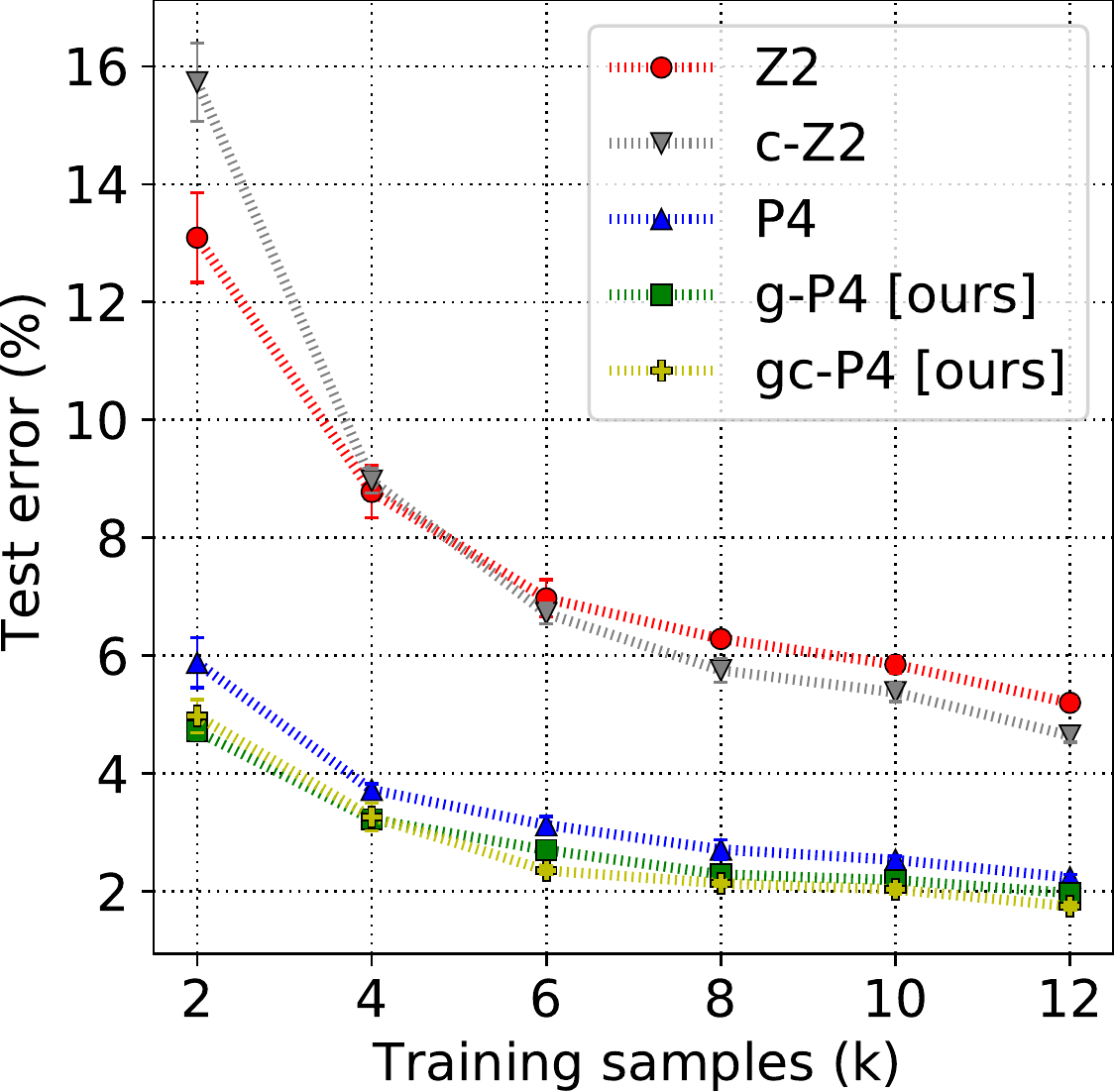}
        \caption{}
        \label{fig:rotmnist_data}
    \end{subfigure}
    \begin{subfigure}[b]{0.49\linewidth}
        \centering
        \includegraphics[width=\textwidth]{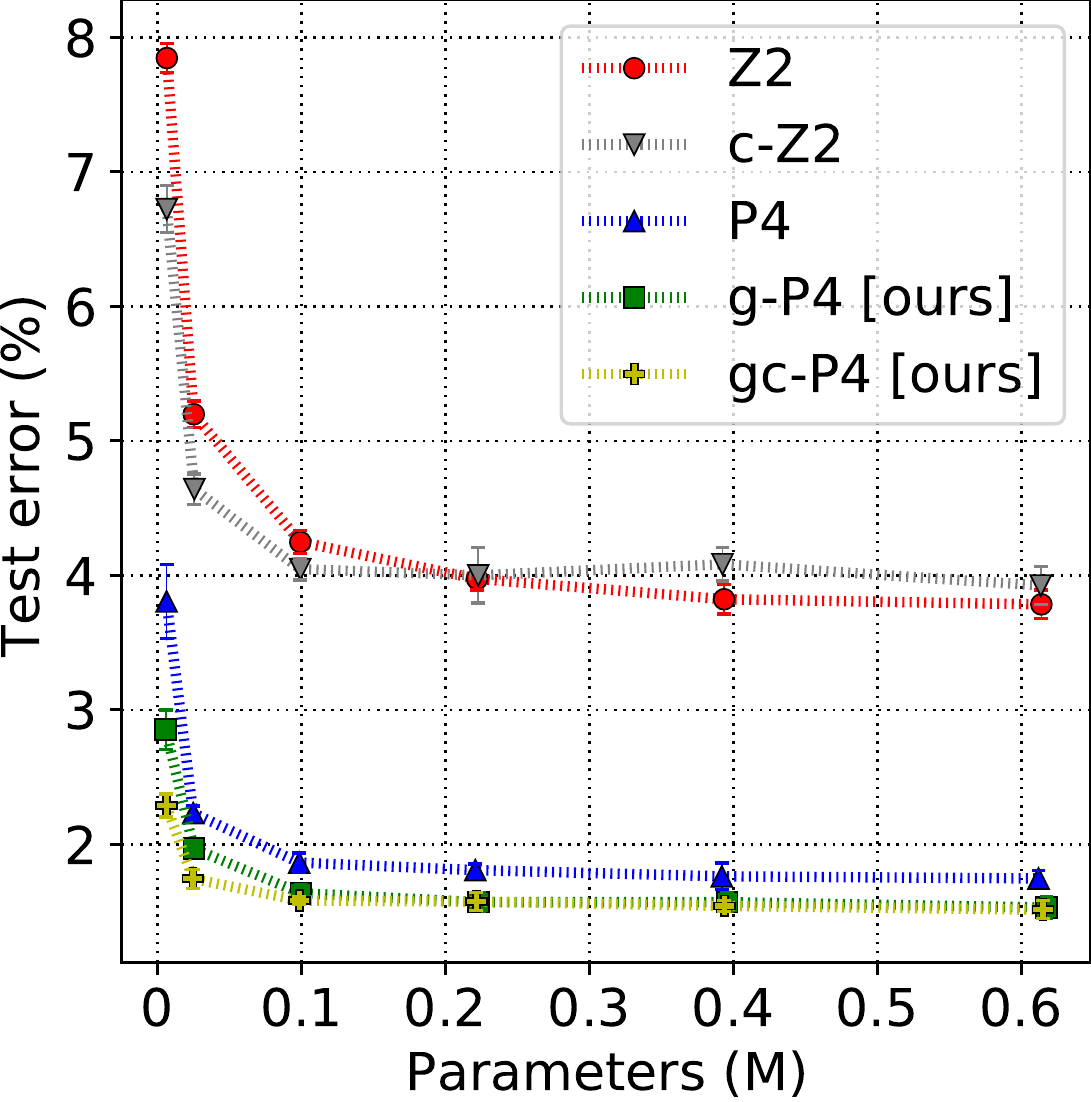}
        \caption{}
        \label{fig:rotmnist_param}
    \end{subfigure}
    \caption{Test error on Rotated MNIST for varying training set (a) and model sizes (b). Architectures with separable GConvs perform consistently better.  }
    \label{fig:rotmnist}
\end{figure}

\tab{rotmnist} shows the test error averaged over 5 runs. Both $g$- and $gc$-P4CNN significantly outperform the regular P4CNN architecture and perform comparably or better than other architectures with a similar parameter count. Both $g$- and $gc$-P4CNN also outperform a depthwise separable version of Z2CNN ($c$-Z2CNN), validating that GConvs are more efficiently decomposable than regular convolutions. Additionally, we evaluate data-efficiency in a reduced data setting. As \fig{rotmnist_data} shows, both $g$- and $gc$-P4CNN consistently outperform  P4CNN. Sharing the same 2D kernel in a GConv filter bank is thus a strong inductive bias and improves the model's sample efficiency. The test error as a function of number of parameters is also shown in \fig{rotmnist_param}. Separable GConvs do better for all model capacities.

\begin{table}[t]
\centering
\caption{Test error on Rotated MNIST - comparison with $z2$ baseline and other $p4$-equivariant methods. $w$ denotes network width. Separable GConv architectures perform better compared to regular GConvs (upper part) and comparable to other equivariant methods (lower part).}
\label{tab:rotmnist}
\begin{tabularx}{\linewidth}{@{}lcccc@{}}
\toprule
\textbf{Network} & \textbf{Test error} & $w$ & \textbf{Param.} & \textbf{MACs} \\ \midrule
Z2CNN~\cite{cohen2016group} & 5.20 $\pm$ 0.110 & 20 & 25.21 k & 2.98 M\\
$c$-Z2CNN & $4.64 \pm 0.126$ & 57 & 25.60 k & 4.14 M \\
P4CNN~\cite{cohen2016group} & 2.23 $\pm$ 0.061 & 10 & 24.81 k & 11.67 M\\
$g$-P4CNN [ours] & 2.60 $\pm$ 0.098 & 10 & 8.91 k & 4.37 M \\
$gc$-P4CNN[ours] & 2.88 $\pm$ 0.169 & 10 & 3.42 k & 1.80 M \\
$g$-P4CNN [ours] & 1.97 $\pm$ 0.044 & 17 & 25.26 k & 12.34 M \\
$gc$-P4CNN [ours] & \textbf{1.74} $\pm$ \textbf{0.070} & 30 & 24.64 k & 13.01 M \\ \midrule
SFCNN~\cite{Weiler_2018_CVPR} & \textbf{0.71} $\pm$ \textbf{0.022} & - & - & - \\
DREN~\cite{li2018deep} & 1.56 & - & 25 k & - \\
H-Net~\cite{Worrall_2017_CVPR} & 1.69 & - & 33 k & - \\
$\alpha$-P4CNN~\cite{romero2020attentive} & $1.70 \pm 0.021$ & 10 & 73.13 k & - \\
$a$-P4CNN~\cite{Romero2020CoAttentive} & 2.06 $\pm$ 0.043 & - & 20.76 k & - \\ \bottomrule
% \multicolumn{4}{l}{\small $^\dagger$same network width as P4CNN.}\\
\end{tabularx}
\end{table}

\subsection{CIFAR 10}
Similarly, we perform a benchmark on the CIFAR10 dataset~\cite{cifar10} using a $p4m$ equivariant version of ResNet44 as detailed in~\cite{cohen2016group}. CIFAR 10+ denotes moderate  data augmentation including random horizontal flips and random translations of up to 4 pixels. Our $gc$-$p4m$-ResNet44 outperforms all other methods using less parameters, as shown in~\tab{cifar10}. Also in a low data regime using only 20\% of the training samples our $gc$-$p4m$ architecture outperforms the regular $p4m$ network with an error rate of 13.43\% vs. 14.20\%.

\begin{table}[t]
\caption{Test error on CIFAR10 - comparison with other $p4m$-equivariant methods. $gc$-$p4m$-ResNet44 performs best.}
\label{tab:cifar10}
\begin{tabularx}{\linewidth}{@{}lYYc@{}}
\toprule
\textbf{Network} & \textbf{CIFAR10} & \textbf{CIFAR10+} & \textbf{Param.} \\ \midrule
ResNet44$^\dagger$~\cite{cohen2016group} &  13.10 & 7.66 & 2.64M \\ 
$p4m$-ResNet44$^\ddag$~\cite{cohen2016group} & 8.06 & 5.78 & 2.62M \\
$\alpha_F$-$p4m$-ResNet44~\cite{romero2020attentive} & 10.82 & 10.12 & 2.70M \\
$a$-$p4m$-ResNet44~\cite{Romero2020CoAttentive} & 9.12 & - & 2.63M \\
$g$-$p4m$-ResNet44 [ours] & 7.60 & 6.09 & 1.78M \\
$gc$-$p4m$-ResNet44 [ours] & \textbf{6.72} & \textbf{5.43} & 1.88M \\ \bottomrule
\multicolumn{4}{@{}l@{}}{\small $^{\dagger\ddag}$ Unable to reproduce results from~\cite{cohen2016group}: 9.45 / 5.61$^\dagger$, 6.46 / 4.94$^\ddag$.}\\
\end{tabularx}
\end{table}

%% file: sections/05_discussion.tex
\section{Discussion}
Our method exploits naturally occurring symmetries in GConvs by explicit sharing of the same filter kernel along the group and input channel dimension using a pointwise and depthwise decomposition. Experiments show that imposing such restriction on the architecture only causes a minor performance drop while allowing to significantly reduce the network parameters. This in turn (i) improves data efficiency and (ii) allows to increase the network width for the same parameter budget resulting in better overall performance. Sharing the spatial kernel over only the group dimension ($g$) proves less effective than additionally sharing over input channels ($gc$) as the latter also efficiently exploits inter-channel correlations in the network. This allows to further increase the network width and thereby its representation power.